\pdfoutput=1

\documentclass[11pt]{article}

\usepackage[]{acl}

\usepackage{times}
\usepackage{latexsym}

\usepackage[T1]{fontenc}

\usepackage[utf8]{inputenc}

\usepackage{microtype}

\usepackage{inconsolata}

\usepackage{multirow}
\usepackage{booktabs}
\usepackage{graphics}
\usepackage{array}
\usepackage[normalem]{ulem}
\useunder{\uline}{\ul}{}
\usepackage{placeins}

\title{Cross-lingual Transfer for Automatic Question Generation \\ by Learning Interrogative Structures in Target Languages}

\author{Seonjeong Hwang$^{\spadesuit}$, Yunsu Kim$^{\clubsuit}$, Gary Geunbae Lee$^{\spadesuit,\diamondsuit}$ \\
         $^{\spadesuit}$ Graduate School of Artificial Intelligence, POSTECH, South Korea \\ 
         $^{\diamondsuit}$ Computer Science and Engineering, POSTECH, South Korea \\
         $^{\clubsuit}$ aiXplain, Inc. Los Gatos, CA, USA \\
         \texttt{seonjeongh@postech.ac.kr}, \texttt{yunsu.kim@aixplain.com}, \texttt{gblee@postech.ac.kr}}

\begin{document}
\maketitle

\begin{abstract}
Automatic question generation (QG) serves a wide range of purposes, such as augmenting question-answering (QA) corpora, enhancing chatbot systems, and developing educational materials.
Despite its importance, most existing datasets predominantly focus on English, resulting in a considerable gap in data availability for other languages.
Cross-lingual transfer for QG (XLT-QG) addresses this limitation by allowing models trained on high-resource language datasets to generate questions in low-resource languages.
In this paper, we propose a simple and efficient XLT-QG method that operates without the need for monolingual, parallel, or labeled data in the target language, utilizing a small language model.
Our model, trained solely on English QA datasets, learns interrogative structures from a limited set of question exemplars, which are then applied to generate questions in the target language.
Experimental results show that our method outperforms several XLT-QG baselines and achieves performance comparable to GPT-3.5-turbo across different languages.
Additionally, the synthetic data generated by our model proves beneficial for training multilingual QA models.
With significantly fewer parameters than large language models and without requiring additional training for target languages, our approach offers an effective solution for QG and QA tasks across various languages\footnote{We release our code and question exemplars used in our experiments at \url{https://github.com/SeonjeongHwang/QuIST}.}.
\end{abstract}

\section{Introduction}

Automatic question generation (QG) aims to generate questions based on a given context.
QG models have been employed not only to augment question-answering (QA) datasets but also to generate educational materials and develop chatbots.
Several QA datasets have been proposed, including SQuAD \cite{rajpurkar2016squad}, HotpotQA \cite{yang2018hotpotqa}, and QuAC \cite{choi2018quac}.
However, the majority of these datasets are in English, resulting in a significant lack of data for other languages.
Moreover, translating English datasets into other languages or creating new QA datasets, despite the availability of similar English datasets, is often inefficient in terms of both time and financial resources.

Recently, researchers have concentrated on cross-lingual transfer (XLT) to address data deficiencies in non-English languages \cite{sherborne2022zero,wu2022zero,vu2022overcoming,deb2023zero,pfeiffer2023mmt5}.
XLT involves deploying models trained on English datasets to other languages when annotated data in the target language is limited or unavailable.

Additionally, in recent years, multilingual large language models (mLLMs), such as GPT-4 \cite{achiam2023gpt}, BLOOM \cite{workshop2022bloom}, and PaLM \cite{chowdhery2023palm}, have exhibited remarkable performance across various natural language generation (NLG) tasks, often achieving high efficacy through zero or few-shot inference.
However, significant cost burdens are associated with utilizing commercial APIs, and employing open-source LLMs requires substantial computational resources.
Previous studies on XLT for QG (XLT-QG) have typically utilized target language data, such as monolingual corpora, source-target parallel corpora, or a limited number of QA examples \cite{kumar2019cross,chi2020cross,shakeri2021towards,wang2021multi,agrawal2023qameleon}.
Nevertheless, incorporating language-specific data during model training can lead to inflexibility in language scalability, necessitating additional training efforts for applications in new languages.

In this paper, we present a simple and efficient XLT-QG method that generates \textbf{Qu}estions by learning \textbf{I}nterrogative \textbf{S}tructures in \textbf{T}arget languages (\textbf{QuIST}).
QuIST comprises two stages: 1) Question Type Classification (QTC) and 2) QG utilizing question exemplars.
We categorize questions into eight types based on English interrogative words, and the QTC model determines the type of question to be generated based on the input context and answer.
Once the question type is identified, it is used to select the corresponding question exemplars for the QG stage.

The QG model generates questions based on a given input context, answer, and question exemplars.
During training with English data, the model learns to identify the interrogative structures specific to each question type from the provided exemplars.
This approach enables the model to generate questions that are not only semantically aligned with the input context and answer but also syntactically similar to the exemplars.
By training exclusively on English data, we ensure that the model can generate questions in other languages without requiring additional training.

In our experiments, we evaluate the performance of QuIST across nine linguistically diverse languages.
Through both automatic and human evaluation, we show that QuIST outperforms various XLT-QG baselines and achieves performance comparable to GPT-3.5-turbo in several languages.
Furthermore, we confirm that synthetic questions generated by QuIST are more effective for training high-performance multilingual QA models than those generated by GPT-3.5-turbo.

Our contributions can be summarized as follows:
\begin{itemize}
    \item We introduce QuIST, a straightforward and efficient XLT-QG method that leverages interrogative structures from question exemplars in target languages during inference.
    \item QuIST exhibits high language scalability, as it can be readily applied to new languages with only a few question exemplars, without requiring additional parameter updates.
    \item Despite utilizing relatively smaller language models, such as mBERT \cite{devlin2018bert} with 110 million parameters and mT5 \cite{xue2021mt5} with 1.2 billion parameters, QuIST generates questions of quality comparable to those produced by GPT-3.5-turbo.
    \item QuIST demonstrates greater effectiveness for data augmentation in multilingual QA compared to other XLT-QG baselines.
\end{itemize}

\section{Cross-lingual Transfer for Automatic Question Generation}

The zero-shot XLT approach--leveraging multilingual pretrained language models (mPLMs) fine-tuned exclusively on English data for target languages--has shown promising performance across various classification tasks \cite{liu2019zero,conneau2019cross,gritta2021xeroalign,wu2022zero,li2023does}.
However, when applied to natural language generation (NLG) tasks, this approach often results in catastrophic forgetting of the target language.
To mitigate this issue,\citet{maurya2021zmbart} proposed fine-tuning only the encoder layers of mPLMs while keeping the word embeddings and all decoder layer parameters frozen.

\begin{figure}[ht]
\centering
\resizebox{\columnwidth}{!}{
\includegraphics{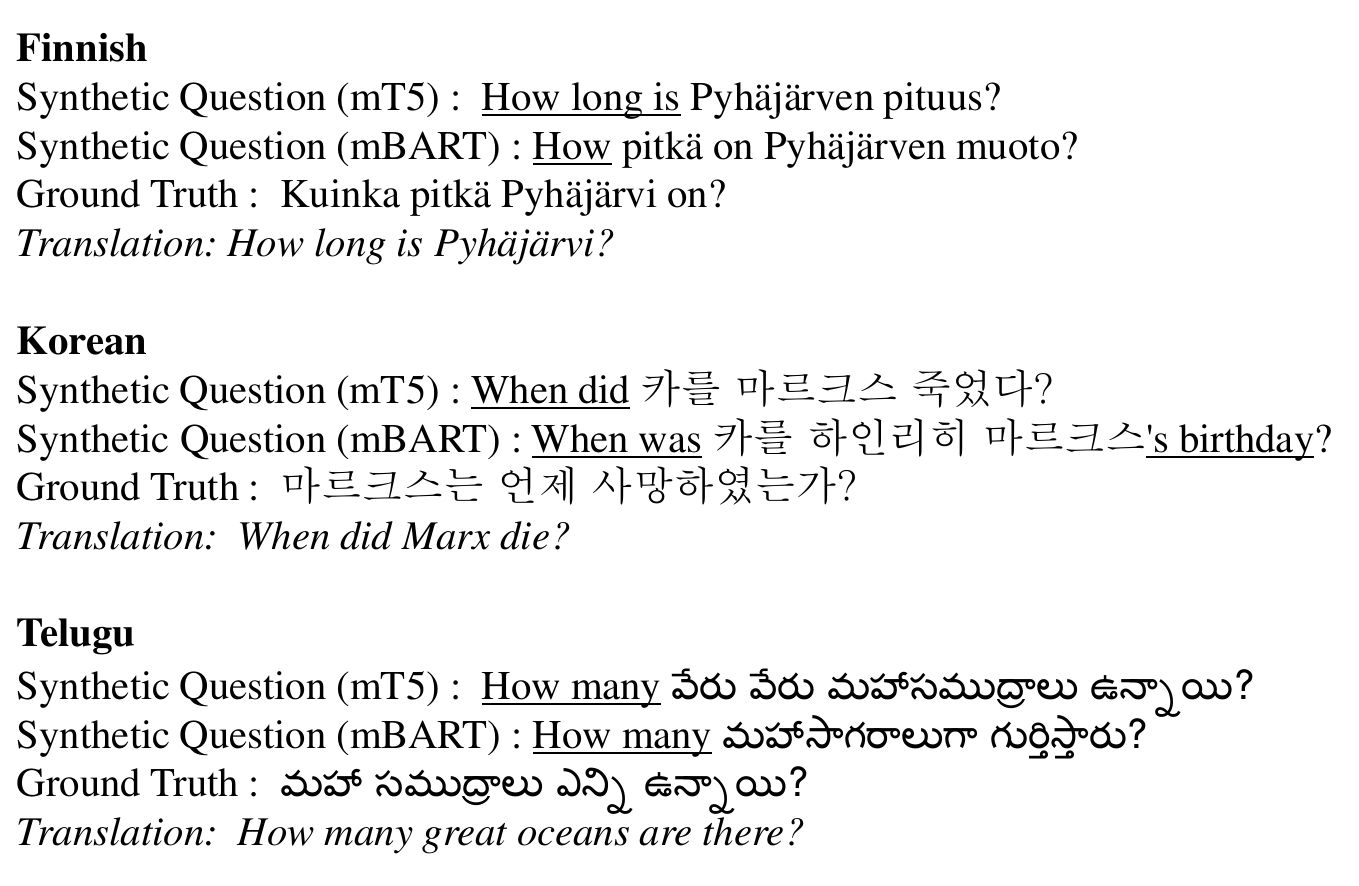}}
\caption{\label{fig:method/code_switch_example} Questions generated by mT5 \cite{xue2021mt5} and mBART \cite{liu2020multilingual} fine-tuned on English QA datasets. The questions often contain English interrogative expressions such as ``How long'' and ``When did.''}
\end{figure}

In our preliminary investigation, we found that this technique did not completely prevent code-switching in XLT-QG, as shown in Figure \ref{fig:method/code_switch_example}.
Specifically, the models struggled to fully grasp interrogative structures in the target language, a phenomenon we refer to as ``interrogative code-switching.''
In this study, we propose a method that enables small mPLMs to learn interrogative structures without relying on target language data during training.

\begin{figure*}[t]
\centering
\resizebox{\textwidth}{!}{
\includegraphics{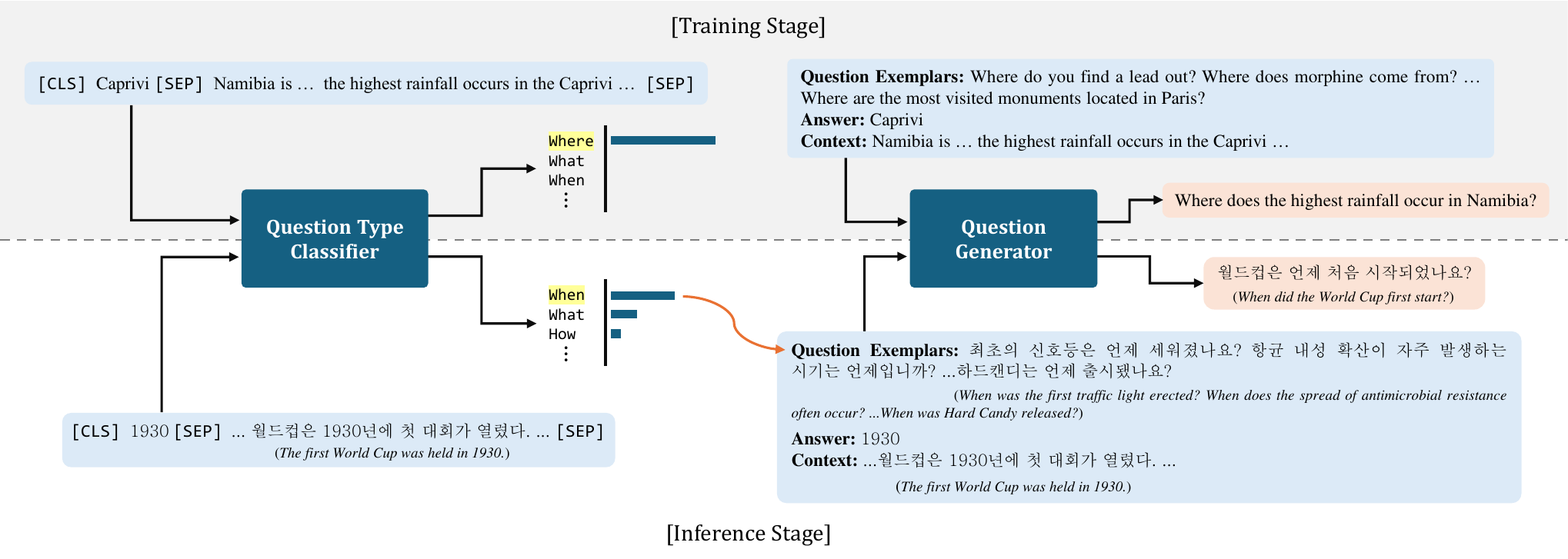}}
\caption{\label{fig:method/architecture} Overview of our proposed method: The QG model generates questions utilizing the question exemplars corresponding to the question type determined by the QTC model.}
\end{figure*}

As illustrated in Figure \ref{fig:method/architecture}, we divide the task into two stages.
In the QTC stage, a classification model identifies the type of question to be generated.
We focus on Wh-questions, categorizing them into eight types based on English interrogative words.
While the type of question is primarily influenced by the type of answer, the model considers both the answer and the context.
This is crucial, as the same answer can result in different types of questions depending on the context.
For example, the number ``911'' could refer to a quantity, year, or proper noun.

The set of question exemplars corresponding to the question type identified by the QTC model is used in the QG stage.
These exemplars are pre-created for each question type and language, as detailed in Section \ref{sec:experiments/data}.
By leveraging the shared interrogative structures among the exemplars, the QG model generates questions using the provided answer and context.
Both the QTC and QG models are trained exclusively on English QA data and can be deployed to new languages without the need for additional training with target language data.

\subsection{\label{sec:QTC} Question Type Classification}

We categorize questions into eight types: \textit{When}, \textit{Where}, \textit{What}, \textit{Which}, \textit{Who}, \textit{Why}, \textit{How$_{way}$}, and \textit{How$_{number}$}\footnote{\textit{How$_{way}$}-inquire about the manner in which something is done, while \textit{How$_{number}$}-questions seek information regarding a degree or specific number.}.
To train the QTC model, we first annotate the question types in the English QA dataset, considering only those questions that fit into one of the eight categories.
Specifically, questions starting with ``how'' are classified as \textit{How$_{way}$} if followed by an auxiliary verb, or as \textit{How$_{number}$} if followed by an adjective or adverb.

In this stage, we apply the zero-shot XLT approach.
We fine-tune mBERT \cite{devlin2018bert} with a feed-forward classification layer using English QA data.
The concatenation of the answer and context, separated by special tokens (i.e., \texttt{[CLS] answer [SEP] context [SEP]}), is fed into the QTC model.
After encoding the input sequence using mBERT, the output hidden vector corresponding to the \texttt{[CLS]} token is passed through a feed-forward layer, followed by the softmax function, to compute probabilities for the eight question types.
We use cross-entropy loss between the predicted and ground-truth labels to update all model parameters.
During inference in target languages, the fine-tuned model predicts the question type by considering the answer and context in those languages.

\subsection{Question Generation with Question Exemplars}

We employ mT5 \cite{xue2021mt5} as the backbone of our QG model, framing the task as a sequence-to-sequence prediction problem. 
The model is trained using the teacher-forcing technique to generate the ground-truth question based on the provided question exemplars, answer, and context. 
During training, the model learns to leverage the syntactic information from the question exemplars to generate questions that are both syntactically correct and semantically appropriate for the given context and answer. 
During inference, the question exemplars corresponding to the question type predicted by the QTC model are input into the QG model, helping it comprehend the interrogative structures of the target language.

\section{Experimental Setup}
In this section, we describe the datasets and baseline models we used in our experiments.
Details regarding the implementation and evaluation metrics are provided in Appendices \ref{sec:appendix:Implementation} and \ref{sec:appendix:metric/auto}, respectively.

\subsection{\label{sec:experiments/data}Data}
\noindent \textbf{QA Datasets} \,
We used SQuAD1.1 \cite{rajpurkar2016squad} as the English QA dataset (\textbf{\textit{C-Q-A$_{en}$}}) to train both the QTC and QG models.
For evaluation, we collected QA examples in nine target languages (\textbf{\textit{C-Q-A$_{tgt}$}}) from multilingual human-annotated QA datasets, including TyDiQA \cite{clark2020tydi}, XQuAD \cite{artetxe2020cross} and MLQA \cite{lewis2020mlqa}.
These datasets consist of \textit{context}--\textit{question}--\textit{answer} triplets, where the answer is a span within the context.
Details about these datasets are provided in Appendix \ref{sec:appendix/data}.

\noindent \textbf{Question Exemplars} \,
The English question exemplars (\textbf{\textit{Q$_{en}$}}) were randomly selected from the questions in the training set of \textit{C-Q-A$_{en}$} after labeling question types as described in Section \ref{sec:QTC}\footnote{In preliminary experiments, we observed that using fixed exemplars was more effective than configuring random exemplars for each training example. A detailed analysis of this finding is provided in Appendix \ref{sec:appendix:Static_Dynamic}.}.
To gather question exemplars in the target languages (\textbf{\textit{Q$_{tgt}$}}) written by native speakers, we utilized the questions from the training set of \textit{C-Q-A$_{tgt}$}.
After translating these questions into English using Google Translation API, we constructed the question exemplars in the same manner as for English.

We experimented with several versions of question exemplars containing different number of questions: \{1, 5, 10, 15\}.
In addition, we randomly sampled each version of the exemplars five times using different random seeds.
Consequently, we trained five distinct QuIST models using different English question exemplars.
During the inference stage, five sets of exemplars for each target language were utilized for evaluation.
As a result, in Section \ref{sec:main results}, we report the average of 25 automatic evaluation results.

\subsection{Baselines}

We compared our QuIST method with several XLT-QG models that share the same backbone, mT5.
All baselines treat the QG task as a sequence-to-sequence prediction, training the models to generate questions based on the concatenation of the input answer and context.

\noindent \textbf{Baseline$_{EncDec}$} \,
This model was simply trained by fine-tuning all parameters of mT5 using \textit{C-Q-A$_{en}$}.
This approach was introduced to examine the effect of training the parameters of the embedding layer and decoder on English data regarding catastrophic forgetting in the target language.

\noindent \textbf{Baseline$_{Enc}$} \,
Unlike Baseline$_{EncDec}$, only parameters of the encoder layers of mT5 were updated for this baseline model.
This training technique was also employed to train QuIST, but the two models differ in whether the question exemplars are utilized.

\noindent \textbf{Baseline$_{Multi}$} \,
Inspired by the method proposed by \citet{shakeri2021towards}, we adopt multi-task fine-tuning, where mT5 simultaneously learns the English QG task and the question denoising task.
The denoising task aims to restore questions with randomly masked tokens and we used \textit{Q$_{tgt}$} with 15 exemplars for each question type (i.e., 120 questions) for a fair comparison with QuIST.
We use this baseline to check whether utilizing a small number of question exemplars during the fine-tuning stage is also effective in XLT-QG.
As this baseline learned language-specific data during training, we constructed different models for each language.

\noindent \textbf{Baseline$_{Adapter}$} \,
We implemented the Adapter-based mT5, which have been recently utilized in XLT for various NLP tasks \cite{pfeiffer2020mad,deb2023zero,pfeiffer2023mmt5,wu2023towards}.
After training language-specific adapters using monolingual corpora\footnote{We extracted 50k raw sentences for each language from the Wikipedia dump (\url{https://dumps.wikimedia.org}) using WikiExtractor (\url{https://github.com/attardi/wikiextractor}), and the language-specific adapters were updated through a text denoising task.}, we trained the task-specific adapters using \textit{C-Q-A$_{en}$}, where the English adapters are incorporated.
While updating each type of adapter, we froze all other model parameters.
In contrast to QuIST, this baseline does not utilize \textit{Q$_{tgt}$}, but instead requires large-scale monolingual corpora in target languages.

\begin{table}[ht]
\centering
\resizebox{\columnwidth}{!}{
\begin{tabular}{l|cc}
\toprule
Model     & \textit{Training}               & \textit{Inference}                                   \\ \midrule
Baseline$_{EncDec}$ & \textit{C-Q-A$_{en}$} & \textit{C-Q-A$_{tgt}$}                        \\
Baseline$_{Enc}$ & \textit{C-Q-A$_{en}$}    & \textit{C-Q-A$_{tgt}$}                        \\
Baseline$_{Multi}$ & \textit{C-Q-A$_{en}$}, \textit{Q$_{tgt}$}  & \textit{C-Q-A$_{tgt}$}                       \\
Baseline$_{Adapter}$ & \textit{C-Q-A$_{en}$}, \textit{S$_{tgt}$}  & \textit{C-Q-A$_{tgt}$}                       \\
QuIST & \textit{C-Q-A$_{en}$}, \textit{Q$_{en}$}    & \textit{C-Q-A$_{tgt}$}, \textit{Q$_{tgt}$} \\ \bottomrule
\end{tabular}}
\caption{\label{tab:experiments/data_for_train_inference} Data utilized by QuIST and baseline models.}
\end{table}

\begin{table*}[t]
\resizebox{\textwidth}{!}{
\begin{tabular}{l|c|ccccccccc|c}
\toprule
Model            & en      & bn              & de              & fi              & hi             & id             & ko             & sw             & te             & zh             & Avg   \\ \midrule
Baseline$_{EncDec}$        & 44.25   & \,\,\,0.72      & 10.11           & 14.48           & \,\,\,2.11     & 13.33          & \,\,\,2.17     & 16.07     & \,\,\,3.92          & 27.63          & 10.06  \\
Baseline$_{Enc}$        & 44.45   & 14.53           & 25.00           & 19.95           & 23.45          & 20.37          & 11.76          & 16.72          & 14.79          & 40.83          & 20.82 \\
Baseline$_{Multi}$        & 41.84   & \,\,\,6.23      & 19.11           & 15.65           & 15.12          & 15.92          & \,\,\,7.92     & 13.65     & \,\,\,8.72          & 30.93          & 14.81 \\
Baseline$_{Adapter}$        & 44.16   & 19.29           & 23.44           & 20.26           & \,\,\,\textbf{31.41}$^{\star}$ & 22.73          & 15.75          & 21.09          & 22.21          & 44.60          & 24.53 \\ \midrule
QuIST$_{1}$      & 43.48   & 14.96           & 25.75           & 27.73           & 21.82          & 23.06          & 11.51          & 20.84          & 10.44          & 42.40          & 22.06 \\
QuIST$_{5}$      & 43.47   & 17.47           & 26.80           & 37.89           & 22.44          & 27.04          & 15.90           & 27.82          & 20.57          & 46.09          & 26.89 \\
QuIST$_{10}$     & 43.40   & \textbf{20.23}  & \textbf{27.08}  & 38.36           & 27.26          & 28.32          & 23.86          & \,\,\,\textbf{31.32}$^{\star}$          & 29.98 & \,\,\,\textbf{47.82}$^{\star}$ & 30.47 \\
QuIST$_{15}$     & 43.08   & 19.07           & 26.84           & \textbf{38.79}  & 27.56          & \textbf{28.36} & \,\,\,\textbf{25.14}$^{\star}$ & 30.59 & \,\,\,\textbf{30.74}$^{\star}$          & 47.71          & \,\,\,\textbf{30.53}$^{\star}$ \\ \midrule
GPT-3.5-turbo$_{zero}$ & 33.98   & 21.30           & 27.76           & 35.55           & 24.84          & 31.18          & 18.56          & 27.90          & 17.31          & 41.67          & 27.34 \\
GPT-3.5-turbo$_{10}$   & 37.63   & \,\,\,21.51$^{\star}$ & \,\,\,29.49$^{\star}$ & \,\,\,39.41$^{\star}$ & 26.60          & \,\,\,32.54$^{\star}$          & 22.28          & 30.12          & 23.13          & 44.47          & 29.95 \\
\bottomrule
\end{tabular}}
\caption{\label{tab:results/main_results}
Automatic evaluation results for the nine target languages.
This table shows the ROUGE-L performance of the models (SP-ROUGE \cite{vu2022overcoming} scores for Chinese). The best scores among mT5-based models are in \textbf{bold} and the highest scores among all models are marked with $\star$. We also report BLEU4 and METEOR scores and standard deviations in Appendix \ref{sec:appendix:Automatic}.}
\end{table*}

Table \ref{tab:experiments/data_for_train_inference} summarizes the datasets utilized by each model during both the training and inference stages. 
As indicated in the table, QuIST, Baseline$_{EncDec}$, and Baseline$_{Enc}$ are exclusively trained on English datasets. 
In contrast, Baseline$_{Multi}$ and Baseline$_{Adapter}$ make use of language-specific data during training. 
Consequently, distinct language-specific models were trained for these two baselines.

\section{\label{sec:main results} Main Results}

\noindent \textbf{Comparison with Baselines} \,
Table \ref{tab:results/main_results} presents the performance of QuIST and the baseline models across nine target languages.
The results show that QuIST$_{15}$, which achieved the highest performance among our models with varying numbers of question exemplars, outperforms several XLT-QG baselines, demonstrating a margin of 6.00 points compared to the most robust baseline, Baseline$_{Adapter}$.
While adapting Baseline$_{Adapter}$ to a new language necessitates training language-specific adapter modules, our model can be readily deployed in new languages without the need for additional training.

QuIST notably outperforms Baseline$_{Enc}$ across all languages.
Note that both models have the same number of trainable parameters during the fine-tuning stage.
These results indicate that exposing the model to interrogative structures during the inference stage significantly enhances its ability to generate questions in the target language.

Despite Baseline$_{Multi}$ learning questions in the target language via the denoising task, it exhibited poor performance, even scoring lower than Baseline$_{Enc}$.
Upon reviewing the generated results of Baseline$_{Multi}$, we frequently observed instances where the questions were unrelated to the input context or answer.
These findings suggest that utilizing a small number of question exemplars during the training stage may lead to overfitting, resulting in a decline in model performance.

\begin{figure*}[t]
\centering
\resizebox{\textwidth}{!}{
\includegraphics{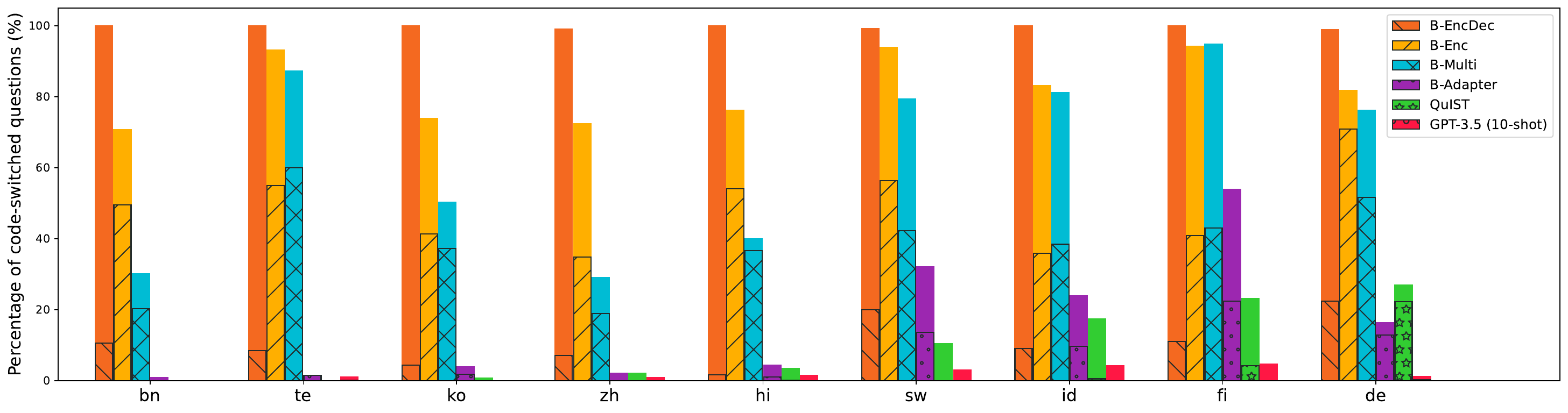}}
\caption{\label{fig:analysis/code_switching_analysis} Percentage of code-switched synthetic questions. The patterned lower section of each bar represents the proportion of questions with only interrogative code-switching, while the full bar indicates the total proportion of all questions involving any form of code-switching.}
\end{figure*}

\begin{table}[ht]
\centering
\small
\begin{tabular}{lccccc}
\toprule
                     & \textit{I}    & \textit{G} & \textit{C} & \textit{A} & \textit{A.M.} \\ \midrule
\multicolumn{6}{c}{bn} \\ \midrule 
Baseline$_{Adapter}$                & \underline{1.10} & 1.60 & \textbf{76.6} & \textbf{76.1} & \textbf{72.3}    \\
QuIST                 & 1.05 & \underline{1.65} & \underline{73.8} & \underline{70.5} & \underline{68.2}    \\
GPT-3.5-turbo$_{10}$               & \textbf{1.69} & \textbf{1.82} & 64.7 & 64.7 & 64.7    \\ \midrule
\multicolumn{6}{c}{de}\\ \midrule
Baseline$_{Adapter}$                & 1.62 & 1.48 & 79.2 & 77.9 & 55.1 \\
QuIST                 & \underline{1.88} & \underline{1.94} & \underline{97.4} & \underline{96.2} & \textbf{96.2} \\
GPT-3.5-turbo$_{10}$               & \textbf{1.96} & \textbf{2.00} & \textbf{100} & \textbf{98.8} & \underline{95.0} \\ \midrule
\multicolumn{6}{c}{fi} \\ \midrule
Baseline$_{Adapter}$                & 0.82 & 1.08 & \textbf{100}  & \textbf{100}  & 73.8    \\
QuIST                 & \underline{1.97} & \underline{1.91} & \textbf{100}  & \textbf{100}  & \textbf{100}    \\
GPT-3.5-turbo$_{10}$               & \textbf{2.00}  & \textbf{1.98} & \textbf{100}  & \textbf{100}  & \underline{98.2}    \\ \midrule
\multicolumn{6}{c}{hi}  \\ \midrule
Baseline$_{Adapter}$                & \underline{1.83} & \underline{1.84} & \underline{31.3} & \underline{32.3} & 20.7  \\
QuIST                 & 1.73 & 1.50 & 28.6 & 26.5 & \textbf{25.7}  \\
GPT-3.5-turbo$_{10}$               & \textbf{1.99} & \textbf{1.96} & \textbf{32.5} & \textbf{32.9} & \underline{24.6}  \\ \midrule 
\multicolumn{6}{c}{id} \\ \midrule
Baseline$_{Adapter}$                & 1.78 & 1.86 & 89.2 & 77.0 & 47.3    \\
QuIST                 & \underline{1.96} & \textbf{2.00} & \textbf{100}  & \underline{98.7} & \underline{97.5}    \\
GPT-3.5-turbo$_{10}$               & \textbf{2.00} & \textbf{2.00} & \textbf{100}  & \textbf{100}  & \textbf{98.8}    \\ \midrule
\multicolumn{6}{c}{sw}  \\ \midrule
Baseline$_{Adapter}$                & 1.36 & 1.73 & 42.4  & 33.9 & 6.8  \\
QuIST                 & \underline{1.94} & \underline{1.82} & \underline{82.5} & \underline{76.3} & \underline{55.0}  \\
GPT-3.5-turbo$_{10}$               & \textbf{2.00} & \textbf{1.95} & \textbf{98.8}   & \textbf{98.8}  & \textbf{96.3}  \\ \bottomrule
\end{tabular}
\caption{\label{tab:analysis/human_evaluation} Human evaluation results.}
\end{table}

\noindent \textbf{Comparison with LLMs} \,
We also compared QuIST and GPT-3.5-turbo, which stands out as a relatively cost-effective option among various commercial LLMs and demonstrates satisfactory results using only a few examples.
We evaluated the performance of GPT-3.5-turbo through zero-shot inference and 10-shot inference, using prompts that included 10 English examples sampled from \textit{C-Q-A$_{en}$}.
The prompt templates we used are provided in Appendix \ref{sec:appendix/gpt}.

According to the results, QuIST$_{15}$ shows higher scores on average than the zero-shot and 10-shot inference of GPT-3.5-turbo.
In detail, our model exhibits better performance in several languages, particularly in Hindi, Korean, Telugu, Swahili, and Chinese.
Additionally, we investigated the few-shot inference of GPT-3.5-turbo that utilized our QTC model and question exemplars. 
The results are reported in Appendix \ref{sec:appendix/qtc-gpt}.

\noindent \textbf{Human Evaluation} \,
We conducted a human evaluation in six languages where QuIST and GPT-3.5-turbo$_{10}$ exhibited similar automatic evaluation results, and we also evaluated the strongest baseline model, Baseline$_{Adapter}$.
We collected a total of 240 questions generated by the three models per language and asked three native speakers to assess the question quality based on five criteria: \textit{Interrogative Sentence} (\textit{I}), \textit{Grammatical Correctness} (\textit{G}), \textit{Clarity} (\textit{C}), \textit{Answerability} (\textit{A}), \textit{Answer-Match} (\textit{A.M.}).
The first two metrics were rated on a scale of {0, 1, 2}, while responses for the remaining categories were binary (yes or no).
More information regarding these criteria is described in Appendix \ref{sec:appendix:metric/human}.

Table \ref{tab:analysis/human_evaluation} presents the majority responses from three raters. 
For the criteria of clarity, answerability, and answer-match, we report the percentage of 'yes' responses.
In German, Finnish, and Indonesian, the questions generated by QuIST and GPT-3.5-turbo$_{10}$ consistently received high scores across all criteria.
Specifically, both models effectively generate questions that align with the given answers, outperforming Baseline$_{Adapter}$.
In contrast, our model achieves lower overall scores in Bengali and Hindi compared to the previously mentioned languages.
However, this performance decline is also observed in GPT-3.5-turbo$_{10}$ and Baseline$_{Adapter}$.

In Swahili, QuIST lagged significantly behind GPT-3.5-turbo$_{10}$ in terms of ``Answerability'' and ``Answer-Match.''
However, given that Baseline$_{Adapter}$ generates questions of significantly lower quality--despite outperforming all other baseline models in automated evaluation--it is noteworthy that our model can generate Swahili questions of acceptable quality without any specific training in the target language.

\section{Analysis}

\begin{table*}[ht]
\centering
\small
\begin{tabular}{lccccccc}
\toprule
Method   & bn             & fi             & id             & ko             & sw             & te             & Avg            \\ \midrule
\textit{English-only}       & 33.63          & 54.05          & 55.75          & 49.03          & 50.30           & 56.40           & 49.86          \\ \midrule
Baseline$_{Enc}$ & {\ul 56.34}    & {\ul 53.71}    & 57.52          & 56.04          & 60.12          & {\ul 68.01}    & 58.62          \\
Baseline$_{Adapter}$ & 54.87          & 50.85          & 58.29          & 52.90           & 59.72          & 64.43          & 56.84          \\
Prompt-tuned PaLM      & 54.57          & \textbf{54.14} & {\ul 59.18}    & {\ul 56.16}    & {\ul 64.00}       & \textbf{69.21} & {\ul 59.54}    \\
GPT-3.5-turbo$_{10}$  & 54.28          & 53.28          & 56.34          & 53.87          & \textbf{64.06} & 64.92          & 57.79          \\
QuIST         & \textbf{59.59} & 53.33          & \textbf{59.53} & \textbf{57.37} & 60.05          & {\ul 68.01}    & \textbf{59.65} \\ \bottomrule
\end{tabular}
\caption{\label{tab:analysis/QA_aug} Exact match scores of multilingual QA models trained on datasets synthesized using different methods.}
\end{table*}

\begin{table*}[ht]
\centering
\small
\begin{tabular}{lccc}
\toprule
Model                                  & \textit{Training} (en) & \textit{Inference} (tgt)       & Avg ROUGE       \\ \midrule
QuIST                                  & Human \& Classified & Human \& Classified            & \textbf{30.53}               \\ \midrule
(1)                                        & Human \& Classified & Translator \& Classified\,\,\,\,\,\,\,  & 27.65               \\
(2)                                        & Human \& Typeless\,\,\,   & Human \& Typeless\,\,\, & 23.59               \\
(3)                                        & $\times$               & Human \& Classified            & 16.96               \\ \midrule
Baseline$_{Enc}$                                  & $\times$             & $\times$             & 20.82               \\ \bottomrule
\end{tabular}
\caption{\label{tab:analysis/ablation} Performance of XLT-QG models using question exemplars in different ways.}
\end{table*}

\subsection{Interrogative Code-switching}
We investigated the frequency of interrogative code-switching occurrence in questions generated by different XLT-QG methods\footnote{We used cld3 (\url{https://github.com/google/cld3}) to identify the languages. If the target language comprised less than 70\% of the generated question, it was classified as code-switching. If the target language accounted for more than 70\% but included English interrogative words, it was classified as interrogative code-switching.}.
As depicted in Figure \ref{fig:analysis/code_switching_analysis}, interrogative code-switching is observed in the majority of questions generated by Baseline$_{EncDec}$.
This phenomenon can be attributed to catastrophic forgetting in target languages, as both the encoder and decoder were fine-tuned using English data.
In Baseline$_{Enc}$, where only the encoder was fine-tuned, this issue is slightly alleviated; nevertheless, more than half of the synthetic questions still exhibit this code-switching problem.

Through the results of Baseline$_{Multi}$, we confirm that interrogative code-switching is alleviated in numerous languages due to the impact of the question denoising task specific to the target language.
Both QuIST and Baseline$_{Adapter}$ prove comparable effectiveness in mitigating interrogative code-switching, surpassing other baseline approaches.
Specifically, our model demonstrates effective in alleviating interrogative code-switching observed in low-resource languages such as Bengali and Swahili.

\subsection{\label{sec:analysis/QA} Data Augmentation for Question Answering}


We explored the potential of QuIST for augmenting training data for multilingual QA models.
Specifically, we compared synthetic data generated by QuIST and baseline models\footnote{The questions were generated based on the context and answer pairs within the synthetic dataset released by \citet{agrawal2023qameleon}.} with the multilingual QA dataset generated by \citet{agrawal2023qameleon}, which used their PaLM-540B model prompt-tuned with five QA examples from target languages.
Table \ref{tab:analysis/QA_aug} presents the average exact match (EM) scores across six languages for the multilingual QA models. 
The training details can be seen in Appendix \ref{sec:appendix:Implementation}.

According to the results, QuIST achieves the best performance, surpassing GPT-3.5-turbo$_{10}$ and prompt-tuned PaLM-540B.
Interestingly, contrary to the findings from the automatic evaluation and interrogative code-switching analysis, Baseline$_{Enc}$ demonstrates greater effectiveness in QA data augmentation compared to Baseline$_{Adapter}$.
In the earlier experiment, over 70\% of the questions generated by Baseline$_{Enc}$ exhibited code-switching issues.
However, unlike Baseline$_{Adapter}$, which depends solely on task-specific adapters for learning the QG task, Baseline$_{Enc}$ leverages all encoder parameters.
This suggests that Baseline$_{Enc}$ may be capable of producing questions with higher semantic quality.

\subsection{Impact of Different Question Exemplars}

We investigated the impact of utilizing different methods for constructing question exemplars compared to our proposed approach.
These approaches were compared to Baseline$_{Enc}$, where only the encoder is fine-tuned on English data, without using additional data from target languages during both training and inference.
Table \ref{tab:analysis/ablation} presents the average ROUGE scores across nine languages.

(1) QuIST utilizes human-written question exemplars in target languages during inference.
In this experiment, we evaluate the model's performance using exemplars translated from English questions via the Google Translation API.
The results show that while machine-translated exemplars improve target language question generation compared to Baseline$_{Enc}$, they are less effective than human-written exemplars.

(2) We conducted training and inference using exemplars that covered all question types to evaluate the effectiveness of type-specific question exemplars.
The exemplars included two instances of each of the eight question types, totaling 16 questions, and the QTC model was not used in this setting.
The results indicate a slight performance improvement compared to Baseline$_{Enc}$; however, this effect is marginal.

(3) We investigated whether input question exemplars during the inference stage are beneficial, even without the training process for generating questions using question exemplars.
The model was trained to generate a question based on the given context and answer without utilizing the question exemplars, similar to Baseline$_{Enc}$, and only used the exemplars in the inference stage.
In this setting, question exemplars in the target language were not helpful, meaning that QuIST learns how to utilize question examples for QG during training.

\subsection{Question Type Classification}

\begin{table}[ht]
\centering
\small
\begin{tabular}{l|c|ccccccccc}
\toprule
Labeling Type   & en    & tgt \\ \midrule
Hard    & 62.92  & 52.86           \\ 
Relaxed & 96.38  & 91.13           \\ \bottomrule
\end{tabular}
\caption{\label{tab:analysis/qtc} Performance of the QTC model.}
\end{table}


To measure the zero-shot inference performance of the QTC model for the target languages, we first annotated the ground-truth question types of the target language QA data.
We translated the questions into English and conducted annotation as detailed in Section \ref{sec:QTC} (i.e., hard labeling).
Table \ref{tab:analysis/qtc} displays the macro F1 scores of the QTC model, measured based on ground-truth labels constructed in two ways.
Since most Wh-questions can be paraphrased into questions beginning with ``what'' and ``which,''\footnote{For example, “How large is the Mupartifad village?” is equivalent to “What is the area of Mupartifad village?”} we also evaluate the QTC performance in a setting where ``what'' and ``which'' are accepted as additional gold labels (i.e., relaxed labeling).
According to the results measured with the relaxed labels, the model correctly classified more than 90\% of questions.
This suggests that the error propagation resulting from misclassification in QTC is minimal throughout the entire pipeline.

\subsection{Case Study}

\begin{figure}[ht]
\centering
\resizebox{\columnwidth}{!}{
\includegraphics{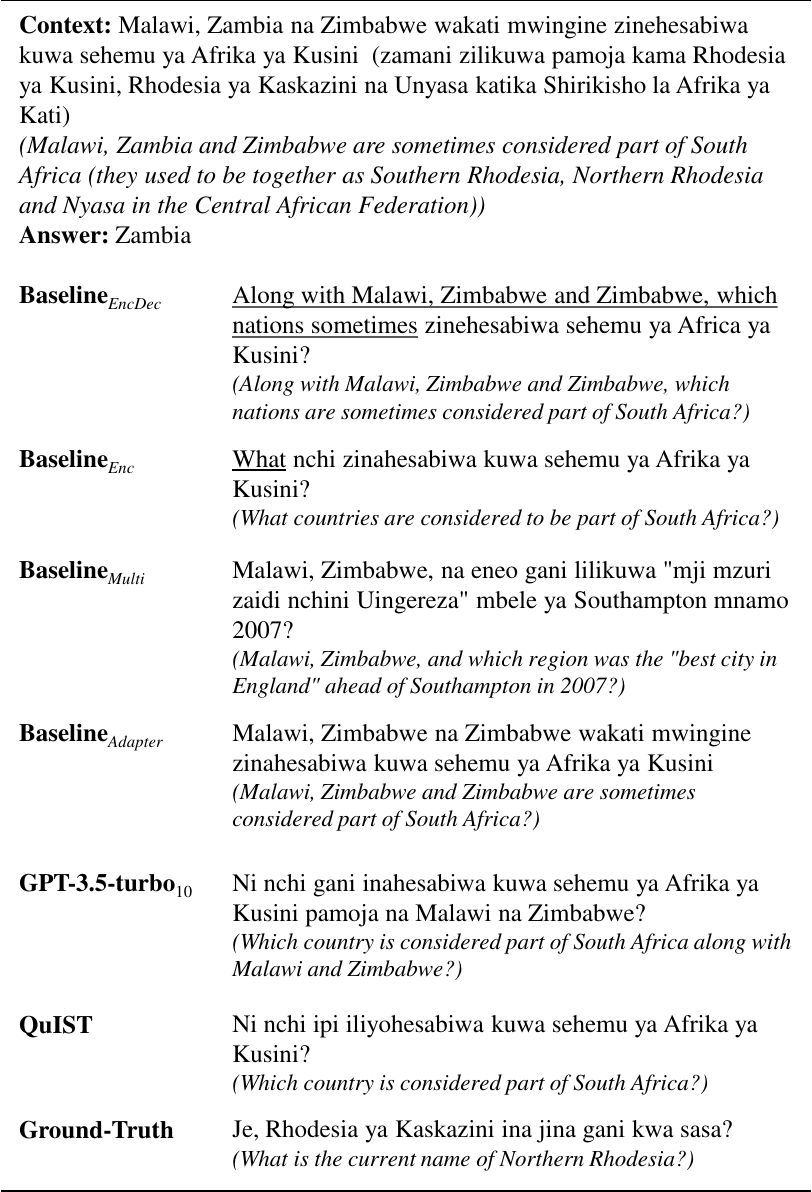}}
\caption{\label{fig:analysis/case_study} Examples of synthetic questions in Swahili.}
\end{figure}

We analyzed the questions generated by the models we used in the experiments, particularly focusing on Swahili, where our model received lower rating than GPT-3.5-turbo$_{10}$ in human evaluation.
In Figure \ref{fig:analysis/case_study}, we can see that the question generated by QuIST is insufficient to explain the given answer, and these incorrect generations resulted in the low ``Answer-Match'' score.
We also note that Baseline$_{EncDec}$ and Baseline$_{Enc}$ encounter code-switching issues, and the question generated by Baseline$_{Multi}$ contains information that is not present in the context.
Furthermore, the question generated by Baseline$_{Adapter}$ was assessed as not being a question, as it is a descriptive sentence ending with a question mark.

\section{Related Work}
Prior work on XLT for NLG tasks has primarily focused on training models with source language datasets while maintaining the ability to generate outputs in the target language. 
For example, \citet{mallinson2020zero} and \citet{chi2020cross} leveraged parallel corpora to improve the alignment between source and target languages, facilitating a more effective transfer of task-related knowledge. 
Similarly, \citet{maurya2021zmbart} enhanced the mPLM model through an auxiliary task closely related to the downstream task, using only monolingual data, and applied it to various NLG tasks in the XLT setting. 
In another approach, \citet{vu2022overcoming} demonstrated that prompt-tuning effectively mitigated catastrophic forgetting of the target language in zero-shot cross-lingual summarization.
More recently, researchers such as \citet{wu2022laft}, \citet{deb2023zero}, and \citet{pfeiffer2023mmt5} have explored methods to separate the acquisition of language-specific knowledge from language-agnostic knowledge, aiming to improve cross-lingual performance.

Unlike most generation tasks that focus on producing declarative sentences, QG involves the additional complexity of generating interrogative sentences designed to elicit specific information. 
While our approach avoids training models using target language data, much of the prior research has relied on such data. 
For instance, \citet{kumar2019cross} utilized a combination of English QA data and a limited amount of target language data. 
In contrast, \citet{shakeri2021towards} trained their model using a denoising task on a question corpus in the target language. 
Additionally, \citet{agrawal2023qameleon} prompt-tuned the PaLM-540B model using five sets of target language QA examples and used the model to synthesize multilingual QA dataset.
Finally, \citet{chi2020cross} adopted a language modeling approach with parallel corpora and restricted the question decoding phase to tokens from the target language vocabulary.

\section{Conclusion}
In this paper, we proposed a simple yet effective XLT-QG approach, where the question generation model is trained solely on an English QA dataset and leverages a small set of target language questions during inference. 
By incorporating question exemplars from target languages, our method enables the model to learn the interrogative structures of those languages, effectively addressing the issue of code-switching.

Experimental results demonstrate that this approach significantly outperforms several XLT-QG baselines and achieves performance comparable to GPT-3.5-turbo across a variety of languages. 
Additionally, we validated the utility of our method’s synthetic data for training multilingual QA models.

A key strength of our method lies in its scalability and parameter efficiency, as it relies exclusively on English QA data during training. 
This enables the seamless extension to new languages without the need for additional parameter updates. 
Moreover, in contrast to LLMs, our approach employs smaller backbone models, offering the advantages of lower deployment costs and reduced computational requirements, making it more accessible for practical use in diverse multilingual settings.

\section{Limitations}
While our model demonstrates strong cross-lingual capabilities, its applicability remains constrained to the languages on which the mPLMs have been trained. 
Although the mT5 model employed in our study was pre-trained on a diverse set of 101 languages, there remain many underrepresented or low-resource languages where the model's performance may be limited.

Another limitation is the instability in model performance, which can vary depending on the configuration of the question exemplars in the target language. 
Some questions generated by the model continue to exhibit code-switching issues.
While this issue may affect the grammatical and linguistic consistency of the outputs, it can be mitigated through the use of a simple rule-based filtering technique. 
Nonetheless, this solution may not entirely eliminate the problem and could require further refinement, particularly in more complex multilingual contexts.

\section*{Acknowledgements}
This research was supported by the MSIT (Ministry of Science and ICT), Korea, under the ITRC (Information Technology Research Center) support program (IITP-2024-RS-2024-00437866) supervised by the IITP (Institute for Information \& Communications Technology Planning \& Evaluation) and also by the Technology Innovation Program (20015007, Development of Digital Therapeutics of Cognitive Behavioral Therapy for treating Panic Disorder) funded By the Ministry of Trade, Industry \& Energy (MOTIE, Korea).

\bibliography{latex/anthology,latex/custom}

\begin{thebibliography}{36}
\expandafter\ifx\csname natexlab\endcsname\relax\def\natexlab#1{#1}\fi

\bibitem[{Achiam et~al.(2023)Achiam, Adler, Agarwal, Ahmad, Akkaya, Aleman, Almeida, Altenschmidt, Altman, Anadkat et~al.}]{achiam2023gpt}
Josh Achiam, Steven Adler, Sandhini Agarwal, Lama Ahmad, Ilge Akkaya, Florencia~Leoni Aleman, Diogo Almeida, Janko Altenschmidt, Sam Altman, Shyamal Anadkat, et~al. 2023.
\newblock Gpt-4 technical report.
\newblock \emph{arXiv preprint arXiv:2303.08774}.

\bibitem[{Agrawal et~al.(2023)Agrawal, Alberti, Huot, Maynez, Ma, Ruder, Ganchev, Das, and Lapata}]{agrawal2023qameleon}
Priyanka Agrawal, Chris Alberti, Fantine Huot, Joshua Maynez, Ji~Ma, Sebastian Ruder, Kuzman Ganchev, Dipanjan Das, and Mirella Lapata. 2023.
\newblock Qameleon: Multilingual qa with only 5 examples.
\newblock \emph{Transactions of the Association for Computational Linguistics}, 11:1754--1771.

\bibitem[{Artetxe et~al.(2020)Artetxe, Ruder, and Yogatama}]{artetxe2020cross}
Mikel Artetxe, Sebastian Ruder, and Dani Yogatama. 2020.
\newblock On the cross-lingual transferability of monolingual representations.
\newblock In \emph{Proceedings of the 58th Annual Meeting of the Association for Computational Linguistics}, pages 4623--4637.

\bibitem[{Banerjee and Lavie(2005)}]{banerjee2005meteor}
Satanjeev Banerjee and Alon Lavie. 2005.
\newblock Meteor: An automatic metric for mt evaluation with improved correlation with human judgments.
\newblock In \emph{Proceedings of the acl workshop on intrinsic and extrinsic evaluation measures for machine translation and/or summarization}, pages 65--72.

\bibitem[{Chi et~al.(2020)Chi, Dong, Wei, Wang, Mao, and Huang}]{chi2020cross}
Zewen Chi, Li~Dong, Furu Wei, Wenhui Wang, Xian-Ling Mao, and Heyan Huang. 2020.
\newblock Cross-lingual natural language generation via pre-training.
\newblock In \emph{Proceedings of the AAAI conference on artificial intelligence}, volume~34, pages 7570--7577.

\bibitem[{Choi et~al.(2018)Choi, He, Iyyer, Yatskar, Yih, Choi, Liang, and Zettlemoyer}]{choi2018quac}
Eunsol Choi, He~He, Mohit Iyyer, Mark Yatskar, Wen-tau Yih, Yejin Choi, Percy Liang, and Luke Zettlemoyer. 2018.
\newblock Quac: Question answering in context.
\newblock In \emph{Proceedings of the 2018 Conference on Empirical Methods in Natural Language Processing}, pages 2174--2184.

\bibitem[{Chowdhery et~al.(2023)Chowdhery, Narang, Devlin, Bosma, Mishra, Roberts, Barham, Chung, Sutton, Gehrmann et~al.}]{chowdhery2023palm}
Aakanksha Chowdhery, Sharan Narang, Jacob Devlin, Maarten Bosma, Gaurav Mishra, Adam Roberts, Paul Barham, Hyung~Won Chung, Charles Sutton, Sebastian Gehrmann, et~al. 2023.
\newblock Palm: Scaling language modeling with pathways.
\newblock \emph{Journal of Machine Learning Research}, 24(240):1--113.

\bibitem[{Clark et~al.(2020)Clark, Choi, Collins, Garrette, Kwiatkowski, Nikolaev, and Palomaki}]{clark2020tydi}
Jonathan~H Clark, Eunsol Choi, Michael Collins, Dan Garrette, Tom Kwiatkowski, Vitaly Nikolaev, and Jennimaria Palomaki. 2020.
\newblock Tydi qa: A benchmark for information-seeking question answering in ty pologically di verse languages.
\newblock \emph{Transactions of the Association for Computational Linguistics}, 8:454--470.

\bibitem[{Conneau and Lample(2019)}]{conneau2019cross}
Alexis Conneau and Guillaume Lample. 2019.
\newblock Cross-lingual language model pretraining.
\newblock \emph{Advances in neural information processing systems}, 32.

\bibitem[{Deb et~al.(2023)Deb, Parab, and Jyothi}]{deb2023zero}
Ujan Deb, Ridayesh Parab, and Preethi Jyothi. 2023.
\newblock Zero-shot cross-lingual transfer with learned projections using unlabeled target-language data.
\newblock In \emph{Proceedings of the 61st Annual Meeting of the Association for Computational Linguistics (Volume 2: Short Papers)}, pages 449--457.

\bibitem[{Devlin et~al.(2018)Devlin, Chang, Lee, and Toutanova}]{devlin2018bert}
Jacob Devlin, Ming-Wei Chang, Kenton Lee, and Kristina Toutanova. 2018.
\newblock Bert: Pre-training of deep bidirectional transformers for language understanding.
\newblock \emph{arXiv preprint arXiv:1810.04805}.

\bibitem[{Gritta and Iacobacci(2021)}]{gritta2021xeroalign}
Milan Gritta and Ignacio Iacobacci. 2021.
\newblock Xeroalign: Zero-shot cross-lingual transformer alignment.
\newblock In \emph{Findings of the Association for Computational Linguistics: ACL-IJCNLP 2021}, pages 371--381.

\bibitem[{Kudo and Richardson(2018)}]{kudo2018sentencepiece}
Taku Kudo and John Richardson. 2018.
\newblock Sentencepiece: A simple and language independent subword tokenizer and detokenizer for neural text processing.
\newblock In \emph{Proceedings of the 2018 Conference on Empirical Methods in Natural Language Processing: System Demonstrations}, pages 66--71.

\bibitem[{Kumar et~al.(2019)Kumar, Joshi, Mukherjee, Ramakrishnan, and Jyothi}]{kumar2019cross}
Vishwajeet Kumar, Nitish Joshi, Arijit Mukherjee, Ganesh Ramakrishnan, and Preethi Jyothi. 2019.
\newblock Cross-lingual training for automatic question generation.
\newblock In \emph{Proceedings of the 57th Annual Meeting of the Association for Computational Linguistics}, pages 4863--4872.

\bibitem[{Lewis et~al.(2020)Lewis, Oguz, Rinott, Riedel, and Schwenk}]{lewis2020mlqa}
Patrick Lewis, Barlas Oguz, Ruty Rinott, Sebastian Riedel, and Holger Schwenk. 2020.
\newblock Mlqa: Evaluating cross-lingual extractive question answering.
\newblock In \emph{Proceedings of the 58th Annual Meeting of the Association for Computational Linguistics}, pages 7315--7330.

\bibitem[{Li and Murray(2023)}]{li2023does}
Tianjian Li and Kenton Murray. 2023.
\newblock Why does zero-shot cross-lingual generation fail? an explanation and a solution.
\newblock In \emph{The 61st Annual Meeting Of The Association For Computational Linguistics}.

\bibitem[{Lin(2004)}]{lin2004rouge}
Chin-Yew Lin. 2004.
\newblock Rouge: A package for automatic evaluation of summaries.
\newblock In \emph{Text summarization branches out}, pages 74--81.

\bibitem[{Liu et~al.(2020)Liu, Gu, Goyal, Li, Edunov, Ghazvininejad, Lewis, and Zettlemoyer}]{liu2020multilingual}
Yinhan Liu, Jiatao Gu, Naman Goyal, Xian Li, Sergey Edunov, Marjan Ghazvininejad, Mike Lewis, and Luke Zettlemoyer. 2020.
\newblock Multilingual denoising pre-training for neural machine translation.
\newblock \emph{Transactions of the Association for Computational Linguistics}, 8:726--742.

\bibitem[{Liu et~al.(2019)Liu, Shin, Xu, Winata, Xu, Madotto, and Fung}]{liu2019zero}
Zihan Liu, Jamin Shin, Yan Xu, Genta~Indra Winata, Peng Xu, Andrea Madotto, and Pascale~Ngan Fung. 2019.
\newblock Zero-shot cross-lingual dialogue systems with transferable latent variables.
\newblock In \emph{EMNLP-IJCNLP 2019-2019 Conference on Empirical Methods in Natural Language Processing and 9th International Joint Conference on Natural Language Processing, Proceedings of the Conference}.

\bibitem[{Loshchilov and Hutter(2018)}]{loshchilov2018decoupled}
Ilya Loshchilov and Frank Hutter. 2018.
\newblock Decoupled weight decay regularization.
\newblock In \emph{International Conference on Learning Representations}.

\bibitem[{Mallinson et~al.(2020)Mallinson, Sennrich, and Lapata}]{mallinson2020zero}
Jonathan Mallinson, Rico Sennrich, and Mirella Lapata. 2020.
\newblock Zero-shot crosslingual sentence simplification.
\newblock In \emph{Proceedings of the 2020 Conference on Empirical Methods in Natural Language Processing (EMNLP)}, pages 5109--5126.

\bibitem[{Maurya et~al.(2021)Maurya, Desarkar, Kano, and Deepshikha}]{maurya2021zmbart}
Kaushal~Kumar Maurya, Maunendra~Sankar Desarkar, Yoshinobu Kano, and Kumari Deepshikha. 2021.
\newblock Zmbart: An unsupervised cross-lingual transfer framework for language generation.
\newblock \emph{arXiv preprint arXiv:2106.01597}.

\bibitem[{Papineni et~al.(2002)Papineni, Roukos, Ward, and Zhu}]{papineni2002bleu}
Kishore Papineni, Salim Roukos, Todd Ward, and Wei-Jing Zhu. 2002.
\newblock Bleu: a method for automatic evaluation of machine translation.
\newblock In \emph{Proceedings of the 40th annual meeting of the Association for Computational Linguistics}, pages 311--318.

\bibitem[{Pfeiffer et~al.(2023)Pfeiffer, Piccinno, Nicosia, Wang, Reid, and Ruder}]{pfeiffer2023mmt5}
Jonas Pfeiffer, Francesco Piccinno, Massimo Nicosia, Xinyi Wang, Machel Reid, and Sebastian Ruder. 2023.
\newblock mmt5: Modular multilingual pre-training solves source language hallucinations.
\newblock In \emph{Findings of the Association for Computational Linguistics: EMNLP 2023}, pages 1978--2008.

\bibitem[{Pfeiffer et~al.(2020)Pfeiffer, Vuli{\'c}, Gurevych, and Ruder}]{pfeiffer2020mad}
Jonas Pfeiffer, Ivan Vuli{\'c}, Iryna Gurevych, and Sebastian Ruder. 2020.
\newblock Mad-x: An adapter-based framework for multi-task cross-lingual transfer.
\newblock In \emph{Proceedings of the 2020 Conference on Empirical Methods in Natural Language Processing (EMNLP)}, pages 7654--7673.

\bibitem[{Rajpurkar et~al.(2016)Rajpurkar, Zhang, Lopyrev, and Liang}]{rajpurkar2016squad}
Pranav Rajpurkar, Jian Zhang, Konstantin Lopyrev, and Percy Liang. 2016.
\newblock Squad: 100,000+ questions for machine comprehension of text.
\newblock In \emph{Proceedings of the 2016 Conference on Empirical Methods in Natural Language Processing}, pages 2383--2392.

\bibitem[{Shakeri et~al.(2021)Shakeri, Constant, Kale, and Xue}]{shakeri2021towards}
Siamak Shakeri, Noah Constant, Mihir Kale, and Linting Xue. 2021.
\newblock Towards zero-shot multilingual synthetic question and answer generation for cross-lingual reading comprehension.
\newblock In \emph{Proceedings of the 14th International Conference on Natural Language Generation}, pages 35--45.

\bibitem[{Sherborne and Lapata(2022)}]{sherborne2022zero}
Tom Sherborne and Mirella Lapata. 2022.
\newblock Zero-shot cross-lingual semantic parsing.
\newblock In \emph{Proceedings of the 60th Annual Meeting of the Association for Computational Linguistics (Volume 1: Long Papers)}, pages 4134--4153.

\bibitem[{Vu et~al.(2022)Vu, Barua, Lester, Cer, Iyyer, and Constant}]{vu2022overcoming}
Tu~Vu, Aditya Barua, Brian Lester, Daniel Cer, Mohit Iyyer, and Noah Constant. 2022.
\newblock Overcoming catastrophic forgetting in zero-shot cross-lingual generation.
\newblock In \emph{Proceedings of the 2022 Conference on Empirical Methods in Natural Language Processing}, pages 9279--9300.

\bibitem[{Wang et~al.(2021)Wang, Yao, Chen, Xu, and Wang}]{wang2021multi}
Bingning Wang, Ting Yao, Weipeng Chen, Jingfang Xu, and Xiaochuan Wang. 2021.
\newblock Multi-lingual question generation with language agnostic language model.
\newblock In \emph{Findings of the Association for Computational Linguistics: ACL-IJCNLP 2021}, pages 2262--2272.

\bibitem[{Workshop et~al.(2022)Workshop, Scao, Fan, Akiki, Pavlick, Ili{\'c}, Hesslow, Castagn{\'e}, Luccioni, Yvon et~al.}]{workshop2022bloom}
BigScience Workshop, Teven~Le Scao, Angela Fan, Christopher Akiki, Ellie Pavlick, Suzana Ili{\'c}, Daniel Hesslow, Roman Castagn{\'e}, Alexandra~Sasha Luccioni, Fran{\c{c}}ois Yvon, et~al. 2022.
\newblock Bloom: A 176b-parameter open-access multilingual language model.
\newblock \emph{arXiv preprint arXiv:2211.05100}.

\bibitem[{Wu et~al.(2022{\natexlab{a}})Wu, Tan, Xu, Liu, Wu, and Song}]{wu2022zero}
Han Wu, Haochen Tan, Kun Xu, Shuqi Liu, Lianwei Wu, and Linqi Song. 2022{\natexlab{a}}.
\newblock Zero-shot cross-lingual conversational semantic role labeling.
\newblock In \emph{Findings of the Association for Computational Linguistics: NAACL 2022}, pages 269--281.

\bibitem[{Wu et~al.(2023)Wu, Zhao, Chang, Shi, Chuang, Chandra, and Juang}]{wu2023towards}
Ting-Wei Wu, Changsheng Zhao, Ernie Chang, Yangyang Shi, Pierce Chuang, Vikas Chandra, and Biing Juang. 2023.
\newblock Towards zero-shot multilingual transfer for code-switched responses.
\newblock In \emph{Proceedings of the 61st Annual Meeting of the Association for Computational Linguistics (Volume 1: Long Papers)}, pages 7551--7563.

\bibitem[{Wu et~al.(2022{\natexlab{b}})Wu, Zheng, Zhou, and Yu}]{wu2022laft}
Xianze Wu, Zaixiang Zheng, Hao Zhou, and Yong Yu. 2022{\natexlab{b}}.
\newblock Laft: Cross-lingual transfer for text generation by language-agnostic finetuning.
\newblock In \emph{Proceedings of the 15th International Conference on Natural Language Generation}, pages 260--266.

\bibitem[{Xue et~al.(2021)Xue, Constant, Roberts, Kale, Al-Rfou, Siddhant, Barua, and Raffel}]{xue2021mt5}
Linting Xue, Noah Constant, Adam Roberts, Mihir Kale, Rami Al-Rfou, Aditya Siddhant, Aditya Barua, and Colin Raffel. 2021.
\newblock mt5: A massively multilingual pre-trained text-to-text transformer.
\newblock In \emph{Proceedings of the 2021 Conference of the North American Chapter of the Association for Computational Linguistics: Human Language Technologies}, pages 483--498.

\bibitem[{Yang et~al.(2018)Yang, Qi, Zhang, Bengio, Cohen, Salakhutdinov, and Manning}]{yang2018hotpotqa}
Zhilin Yang, Peng Qi, Saizheng Zhang, Yoshua Bengio, William Cohen, Ruslan Salakhutdinov, and Christopher~D Manning. 2018.
\newblock Hotpotqa: A dataset for diverse, explainable multi-hop question answering.
\newblock In \emph{Proceedings of the 2018 Conference on Empirical Methods in Natural Language Processing}. Association for Computational Linguistics.

\end{thebibliography}

\appendix

\section{\label{sec:appendix:Static_Dynamic} Static and Dynamic Exemplars}
Since gathering sufficient question samples in the target languages is challenging, we used fixed question exemplars during inference. 
In contrast, English question exemplars can be easily sourced from QA datasets. 
Thus, we experimented with two approaches for creating question exemplars to train the QG model: (1) Static exemplars, which use fixed exemplars across all training examples, and (2) Dynamic exemplars, which are sampled from the English QA dataset for each training example.

\begin{table}[ht]
\centering
\resizebox{0.8\columnwidth}{!}{
\begin{tabular}{ccc}
\toprule
Language & Dynamic                   & Static                    \\ \midrule
bn       & \textbf{20.40} $\pm$ 0.40 & 20.13 $\pm$ 0.71          \\
de       & 26.53 $\pm$ 0.19          & \textbf{26.84} $\pm$ 0.26 \\
fi       & 35.91 $\pm$ 1.35          & \textbf{43.09} $\pm$ 2.18 \\
hi       & 26.97 $\pm$ 0.44          & \textbf{27.72} $\pm$ 0.28 \\
id       & 27.42 $\pm$ 2.04          & \textbf{29.96} $\pm$ 2.22 \\
ko       & 23.35 $\pm$ 0.37          & \textbf{27.01} $\pm$ 0.83 \\
sw       & 27.57 $\pm$ 0.86          & \textbf{32.01} $\pm$ 1.09 \\
te       & 27.64 $\pm$ 0.99          & \textbf{32.96} $\pm$ 1.17 \\
zh       & 47.29 $\pm$ 0.22          & \textbf{47.64} $\pm$ 0.26 \\ \midrule
AVG      & 29.23                     & \textbf{31.93}            \\ \bottomrule
\end{tabular}}
\caption{\label{tab:static_dynamic} Comparison of models using dynamic and static exemplars during training. We report SP-ROUGE scores for Chinese and ROUGE-L scores for other languages. The scores for the static setting are based on the English exemplars, representing median performance.}
\end{table}

As shown in Table \ref{tab:static_dynamic}, both approaches demonstrate effective performance in target languages compared to the existing XLT-QG baseline models (Table \ref{tab:results/main_results}).
However, the static exemplar method achieves better overall performance across various languages.
During training, our model generates questions by leveraging the syntactic information from the exemplars while utilizing the semantic information from the input context and answer.
We hypothesize that the model trained with static exemplars was better able to focus on the syntactic structures of the example questions, leading to improved performance.
Consequently, we utilized static exemplars in all our experiments.

\section{\label{sec:appendix:Implementation} Implementation Details}
We utilized a single NVIDIA Tesla A100-80GB GPU for model training. 
The QTC and QG models were initialized using \texttt{bert-base-multilingual-cased} with 110M parameters and \texttt{google/mt5-large} with 1.2B parameters, sourced from HuggingFace\footnote{\url{https://huggingface.co}}. 
Training was conducted employing stochastic gradient descent with the AdamW optimizer \cite{loshchilov2018decoupled} coupled with a linear learning rate scheduler encompassing 1000 warm-up steps. 
Batch sizes and learning rates were set as (8, 1e-5) and (16, 5e-5) for QTC and QG, respectively. 
Training ceased upon optimization of the models on the validation set.

Due to variations in the number of examples across different question types, we employed data upsampling based on the type with the highest number of examples for training the QTC model. 
During the inference stage, we determined the question type with the highest predicted probability from the QTC model and generated questions using the beam search algorithm with a beam size of 4.

To train multilingual QA models in Section \ref{sec:analysis/QA}, we adopted the methodologies used by \citet{agrawal2023qameleon}. 
Each QA model underwent training using a combination of English data sourced from the TyDiQA training set and synthetic data for all languages, generated by each XLT-QG model. 
Given the unavailability of the TyDiQA test set, we evaluated the validation performance instead. 
The backbone of the QA model consisted of \texttt{google/mt5-xl} with 3.7B parameters, fine-tuned with a learning rate of 2e-4 and a batch size of 64. 
We selected the model checkpoint yielding the highest EM score for each language, following the strategy of \citet{agrawal2023qameleon}, and reported the average scores obtained from utilizing three different random seeds.

\section{\label{sec:appendix:metric} Metric}
\subsection{\label{sec:appendix:metric/auto}Automatic Evaluation}
In accordance with previous studies on QG, we use BLEU4 \cite{papineni2002bleu}, METEOR \cite{banerjee2005meteor}, ROUGE-L \cite{lin2004rouge} as automatic evaluation metrics.
These metrics measure the n-gram similarity between model predictions and references.
However, these evaluation metrics are not suitable for Chinese (zh), where words are not separated by white space.
Therefore, we additionally used SP-ROUGE \cite{vu2022overcoming} that using SentencePiece sub-word tokenization \cite{kudo2018sentencepiece}.

\subsection{\label{sec:appendix:metric/human} Human Evaluation}

We enlisted three native speakers for each language via Upwork\footnote{\url{https://www.upwork.com}} to evaluate the quality of our synthetic questions.
The questions were rated based on five criteria:
\begin{itemize}
    \item \textit{Interrogative Sentence} evaluates whether the question has an interrogative structure. \\
        \textbf{0}: This is not a question. \\
        \textbf{1}: This is a question, but it doesn't have the typical structure of an interrogative sentence. \\
        \textbf{2}: This is a natural interrogative structure. 
    \item \textit{Grammatical Correctness} evaluates the grammatical accuracy of the question. \\
        \textbf{0}: Numerous grammatical errors make the question unacceptable. \\
        \textbf{1}: Some errors exist but do not hinder understanding of the question. \\
        \textbf{2}: The question is grammatically correct.
    \item \textit{Clarity} determines whether the question is clear and easily understandable given the context. Answer \textbf{yes} or \textbf{no}.
    \item \textit{Answerability} determines whether the question can be answered using information from the context. Answer \textbf{yes} or \textbf{no}.
    \item \textit{Answer-Match} determines whether the input answer could be a valid answer to the question considering the content of the provided context. Answer \textbf{yes} or \textbf{no}.
\end{itemize}
If a score of ``0'' is assigned to the \textit{Interrogative Sentence} category, evaluations for the remaining categories did not conducted. 
Additionally, if a score of 0 is rated in \textit{Grammatical Correctness}, or if ``no'' is selected for \textit{Clarity}, \textit{Answerability}, or \textit{Answer-Match} categories, subsequent evaluations can not be carried out. 
Therefore, in this case, the lowest scores were assigned for these criteria.

\section{\label{sec:appendix/data} Data Usage}

We used SQuAD1.1 \cite{rajpurkar2016squad} as the English QA data \textit{C-Q-A$_{en}$} for training our models. 
As only training and validation sets are publicly available, we partitioned the training set and employed a portion of the examples for validation purposes. 
The original validation set served as our test set. 
The training, validation, and test sets comprised 79,321, 8,283, and 1,190 examples, respectively. 
Furthermore, the distribution of examples by question type is summarized in Table \ref{tab:appendix/data/QTC}.

\begin{table}[ht]
\centering
\small
\begin{tabular}{cc}
\toprule
Question Type & \# Examples \\ \midrule
\textit{What}          & 33,777      \\
\textit{Who}           & \,\,\,7,951       \\
\textit{How$_{number}$} & \,\,\,5,657      \\
\textit{When}          & \,\,\,4,780       \\
\textit{Which}         & \,\,\,3,931       \\
\textit{Where}         & \,\,\,2,953       \\
\textit{How$_{way}$}   & \,\,\,1,600       \\
\textit{Why}           & \,\,\,1,054        \\
\bottomrule
\end{tabular}
\caption{\label{tab:appendix/data/QTC} Number of examples by question type in training set of \textit{C-Q-A$_{en}$}.}
\end{table}

\begin{table}[ht]
\centering
\resizebox{0.8\columnwidth}{!}{
\begin{tabular}{cccc}
\toprule
\multirow{2}{*}{Language} & \multirow{2}{*}{Code} & \multicolumn{2}{c}{\# Examples} \\ 
                          &                       & Train          & Test           \\ \midrule
Bengali                   & bn                    & 2,390          & \,\,\,\,\,113            \\
Chinese                   & zh                    & 5,137          & 1,190          \\
German                    & de                    & 4,517          & 1,190          \\
Finnish                   & fi                    & 6,855          & \,\,\,\,\,782            \\
Hindi                     & hi                    & 4,918          & 1,190          \\
Indonesian                & id                    & 5,702          & \,\,\,\,\,565            \\
Korean                    & ko                    & 1,625          & \,\,\,\,\,276            \\
Swahili                   & sw                    & 2,755          & \,\,\,\,\,499            \\
Telugu                    & te                    & 5,563          & \,\,\,\,\,669            \\ \bottomrule
\end{tabular}}
\caption{\label{tab:experiments/data_statistics} Language codes and the number of examples in \textit{C-Q-A$_{tgt}$} dataset. In our method, only a small portion of the training examples are used as question exemplars.}
\end{table}

Table \ref{tab:experiments/data_statistics} presents the statistics of target language QA data \textit{C-Q-A$_{tgt}$} utilized by our models during inference. 
Note that training examples were solely employed for sampling question exemplars \textit{Q$_{tgt}$}. 
Test examples in Chinese, German, and Hindi were collected from the XQuAD \cite{artetxe2020cross} test set, whereas training examples were sourced from the MLQA \cite{lewis2020mlqa} validation set, as XQuAD does not provide a training set for the target languages.
Training and test examples in other languages were obtained from TyDiQA \cite{clark2020tydi}.

\section{\label{sec:appendix/gpt} Prompt Template for GPT-3.5-turbo}

\begin{figure}[ht]
\centering
\resizebox{\columnwidth}{!}{
\includegraphics{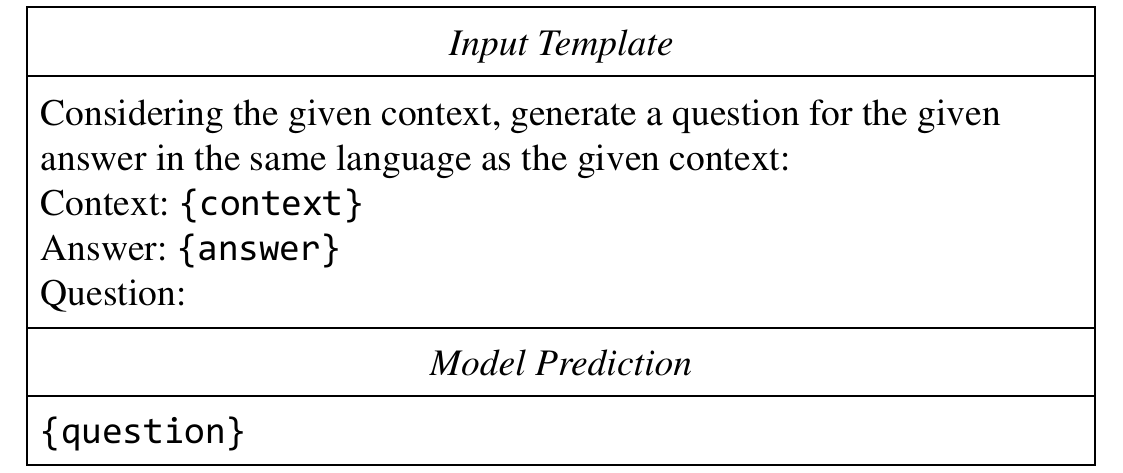}}
\caption{\label{fig:appendix:zero-shot} The input and output template for zero-shot inference of GPT-3.5-turbo.}
\end{figure}

\begin{figure}[ht]
\centering
\resizebox{\columnwidth}{!}{
\includegraphics{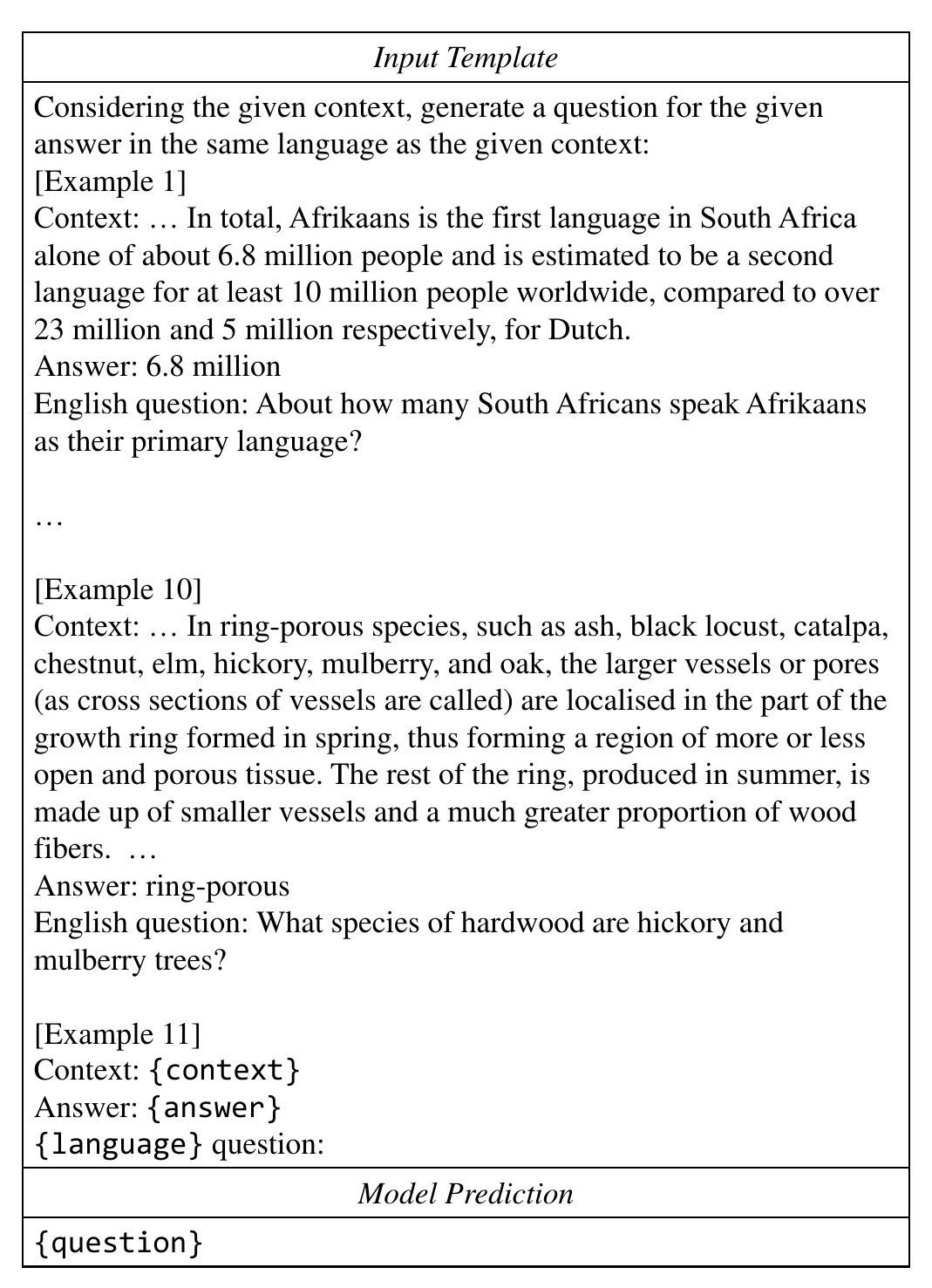}}
\caption{\label{fig:appendix:few-shot} The input and output template for 10-shot inference of GPT-3.5-turbo.}
\end{figure}

We evaluated the zero-shot and few-shot performance of \texttt{gpt-3.5-turbo-0125} model.
We extracted sets with different numbers of examples: 1, 3, 5, and 10, from \textit{C-Q-A$_{en}$} to employ for few-shot inference.
In addition, we used five versions of each set, varying the random seed.
Based on the English validation set, we determined the optimal number of examples (see Table \ref{tab:appendix:gpt_validation}), and used the set with the median performance as the component in the few-shot prompt. 
Subsequently, we conducted zero-shot and 10-shot inference for various languages using the prompts described in Figure \ref{fig:appendix:zero-shot} and \ref{fig:appendix:few-shot}, respectively. 

\begin{table}[ht]
\resizebox{\columnwidth}{!}{
\begin{tabular}{cc|ccc}
\toprule
\multicolumn{2}{c|}{Prompt Type}                    & BLEU-4                & METEOR                & ROUGE-L               \\ \midrule
\multicolumn{2}{c|}{Zero-shot}                      & 15.01                 & 53.28                 & 40.32                 \\ \midrule
\multicolumn{1}{c|}{\multirow{4}{*}{Few-shot}} & 1  & 17.58 $\pm$ 3.04          & 52.99 $\pm$ 0.80          & 40.20 $\pm$ 3.11          \\
\multicolumn{1}{l|}{}                          & 3  & 18.28 $\pm$ 1.82          & 53.43 $\pm$ 1.01          & 41.10 $\pm$ 1.71          \\
\multicolumn{1}{l|}{}                          & 5  & 19.09 $\pm$ 0.85          & 54.02 $\pm$ 1.11          & 41.40 $\pm$ 1.27          \\
\multicolumn{1}{l|}{}                          & 10 & \textbf{19.42} $\pm$ 1.02 & \textbf{54.37} $\pm$ 0.69 & \textbf{42.10} $\pm$ 1.01 \\ \bottomrule
\end{tabular}}
\caption{\label{tab:appendix:gpt_validation} Performance of GPT-3.5-turbo on the SQuAD1.1 validation set. We report the mean and standard deviation of the few-shot inference results.}
\end{table}

Additionally, we empirically observed that specifying the language of the questions to be generated is essential for effective few-shot inference. 
Even when the input context and answer are in non-English languages, the model frequently generated English questions when the language to be generated was not specified.

\section{\label{sec:appendix:Automatic} Automatic Evaluation Results}
Table \ref{tab:appendix:bleu4}, \ref{tab:appendix:meteor}, and \ref{tab:appendix:rouge_l} show detailed results for the experiments in Section \ref{sec:main results}.

\section{\label{sec:appendix/qtc-gpt} GPT-3.5-turbo few-shot Inference with Question Type Classification}

We additionally investigated whether the QTC model and question exemplars are beneficial for few-shot inference of GPT-3.5-turbo.
In this experiment, we utilized the exemplar set that exhibited the best performance for each language in our method.
We supplemented these exemplars with the statement ``The followings are examples of \texttt{language} questions:'' placed before the prompt in Figure \ref{fig:appendix:few-shot}.
According to the results in Table \ref{tab:appendix/qtc-gpt}, leveraging the QTC model and question exemplars leads to particularly improved performance in low-resource languages such as Bengali, Telugu, and Swahili.

\begin{table*}[ht]
\resizebox{\textwidth}{!}{
\begin{tabular}{l|c|cccccccc|c}
\toprule
Model                  & en               & bn              & de               & fi               & hi              & id                      & ko                    & sw                    & te                    & Avg        \\ \midrule
Baseline$_{EncDec}$    & 23.45            & 0.00            & \,\,\,3.62       & 2.91             & \,\,\,0.35      & \,\,\,5.59              & 0.00                  & 4.46                  & 0.97                  & 2.24       \\
Baseline$_{Enc}$       & 23.72            & 5.64            & 13.57            & 6.27             & 10.01           & 10.11                   & 4.38                  & 5.80                  & 3.64                  & 7.43       \\
Baseline$_{Multi}$     & 23.45            & 2.04            & \,\,\,9.38       & 3.17             & \,\,\,3.63      & \,\,\,6.46              & 1.85                  & 2.35                  & 1.77                  & 3.83       \\
Baseline$_{Adapter}$   & 21.79            & 6.96            & 11.34            & 5.57             & 12.28           & \,\,\,9.10              & 4.41                  & 6.38                  & 6.41                  & 7.81       \\ \midrule
QuIST$_{1}$            & 22.32 $\pm$ 0.06 & 5.18 $\pm$ 0.72 & 12.97 $\pm$ 0.39 & 13.02 $\pm$ 2.04 & 7.78 $\pm$ 1.31 & 12.81 $\pm$ 0.88  & \,\,\,2.54 $\pm$ 1.50 & \,\,\,8.24 $\pm$ 2.18 & \,\,\,3.41 $\pm$ 1.05 & \,\,\,8.24 \\
QuIST$_{5}$            & 22.20 $\pm$ 0.13 & 6.62 $\pm$ 0.97 & 13.43 $\pm$ 0.30 & 20.50 $\pm$ 1.54 & 7.84 $\pm$ 1.19 & 14.78 $\pm$ 0.79  & \,\,\,5.57 $\pm$ 4.21 & 15.07 $\pm$ 2.87      & \,\,\,9.38 $\pm$ 4.27 & 11.65      \\
QuIST$_{10}$           & 22.17 $\pm$ 0.14 & 7.88 $\pm$ 0.70 & 13.43 $\pm$ 0.26 & 19.71 $\pm$ 2.57 & 9.44 $\pm$ 0.75 & 15.59 $\pm$ 0.94  & 10.87 $\pm$ 1.97      & 18.29 $\pm$ 1.39      & 13.19 $\pm$ 3.84      & 13.55      \\
QuIST$_{15}$           & 21.90 $\pm$ 0.10 & 7.20 $\pm$ 0.75 & 13.49 $\pm$ 0.27 & 20.46 $\pm$ 2.52 & 9.15 $\pm$ 0.38 & 15.34 $\pm$ 1.38       & 11.26 $\pm$ 1.07      & 17.34 $\pm$ 1.37      & 13.83 $\pm$ 3.05      & 13.51      \\ \midrule
GPT-3.5-turbo$_{zero}$ & 12.27            & 7.76            & 11.53            & 11.84            & 7.53            & 11.25                   & 5.40                  & 10.90                 & 4.59                  & \,\,\,8.85 \\
GPT-3.5-turbo$_{10}$   & 15.50            & 7.77            & 12.40            & 15.45            & 7.30            & 12.84                   & 7.82                  & 11.55                 & 5.30                  & 10.05      \\ \bottomrule
\end{tabular}}
\caption{\label{tab:appendix:bleu4} Automatic evaluation results using BLEU4.}
\end{table*}

\begin{table*}[ht]
\resizebox{\textwidth}{!}{
\begin{tabular}{l|c|cccccccc|c}
\toprule
Model                  & en               & bn               & de               & fi               & hi               & id               & ko               & sw               & te               & Avg   \\ \midrule
Baseline$_{EncDec}$    & 50.98            & \,\,\,6.95       & 16.09            & 21.72            & \,\,\,6.29       & 25.25            & 10.38            & 22.85            & 13.06            & 15.32 \\
Baseline$_{Enc}$       & 50.68            & 22.21            & 31.23            & 27.92            & 27.73            & 35.10            & 17.78            & 25.79            & 23.05            & 26.35 \\
Baseline$_{Multi}$     & 50.99            & 11.68            & 24.88            & 23.16            & 18.24            & 28.36            & 14.99            & 18.76            & 16.93            & 19.63 \\
Baseline$_{Adapter}$   & 48.11            & 24.96            & 31.30            & 29.47            & 33.47            & 36.57            & 16.04            & 28.03            & 23.50            & 27.92 \\ \midrule
QuIST$_{1}$            & 48.67 $\pm$ 0.12 & 21.69 $\pm$ 1.60 & 30.57 $\pm$ 0.74 & 34.14 $\pm$ 2.87 & 25.15 $\pm$ 3.01 & 36.99 $\pm$ 1.74 & 17.26 $\pm$ 1.01 & 31.73 $\pm$ 3.01 & 22.09 $\pm$ 1.69 & 27.45 \\
QuIST$_{5}$            & 48.56 $\pm$ 0.14 & 23.66 $\pm$ 1.23 & 31.39 $\pm$ 0.40 & 41.57 $\pm$ 1.60 & 25.36 $\pm$ 2.69 & 40.85 $\pm$ 1.07 & 19.94 $\pm$ 4.24 & 40.19 $\pm$ 3.25 & 27.59 $\pm$ 3.99 & 31.32 \\
QuIST$_{10}$           & 48.51 $\pm$ 0.19 & 25.22 $\pm$ 1.28 & 31.33 $\pm$ 0.40 & 41.78 $\pm$ 1.59 & 28.85 $\pm$ 1.60 & 41.66 $\pm$ 1.96 & 24.74 $\pm$ 3.52 & 43.89 $\pm$ 1.31 & 30.62 $\pm$ 3.18 & 33.51 \\
QuIST$_{15}$           & 48.22 $\pm$ 0.12 & 24.49 $\pm$ 1.45 & 31.43 $\pm$ 0.47 & 42.38 $\pm$ 2.64 & 29.51 $\pm$ 0.79 & 42.38 $\pm$ 2.39 & 27.65 $\pm$ 2.47 & 43.15 $\pm$ 1.80 & 32.65 $\pm$ 1.77 & 34.21 \\ \midrule
GPT-3.5-turbo$_{zero}$ & 47.61            & 27.08            & 35.50            & 41.48            & 28.84            & 45.81            & 23.19            & 41.10            & 24.16            & 33.40 \\
GPT-3.5-turbo$_{10}$   & 49.29            & 26.82            & 37.43            & 44.72            & 30.16            & 47.05            & 27.98            & 40.96            & 27.49            & 35.33 \\ \bottomrule
\end{tabular}}
\caption{\label{tab:appendix:meteor}Automatic evaluation results using METEOR.}
\end{table*}

\begin{table*}[ht]
\resizebox{\textwidth}{!}{
\begin{tabular}{l|c|ccccccccc|c}
\toprule
Model                  & en               & bn               & de               & fi               & hi               & id               & ko               & sw               & te               & zh               & Avg   \\ \midrule
Baseline$_{EncDec}$    & 44.25            & \,\,\,0.72       & 10.11            & 14.48            & \,\,\,2.11       & 13.33            & \,\,\,2.17       & 16.07            & \,\,\,3.92            & 27.63            & 10.06 \\
Baseline$_{Enc}$       & 44.45            & 14.53            & 25.00            & 19.95            & 23.45            & 20.37            & 11.76            & 16.72            & 14.79            & 40.83            & 20.82 \\
Baseline$_{Multi}$     & 41.84            & \,\,\,6.23       & 19.11            & 15.65            & 15.12            & 15.92            & \,\,\,7.92       & 13.65            & \,\,\,8.72       & 30.93            & 14.81 \\
Baseline$_{Adapter}$   & 44.16            & 19.29            & 23.44            & 20.26            & 31.41            & 22.73            & 15.75            & 21.09            & 22.21            & 44.60            & 24.53 \\ \midrule
QuIST$_{1}$            & 43.48 $\pm$ 0.04 & 14.96 $\pm$ 2.05 & 25.75 $\pm$ 0.87 & 27.73 $\pm$ 3.87 & 21.82 $\pm$ 3.50 & 23.06 $\pm$ 2.14 & 11.51 $\pm$ 1.07 & 20.84 $\pm$ 2.44 & 10.44 $\pm$ 3.22 & 42.40 $\pm$ 2.32 & 22.06 \\
QuIST$_{5}$            & 43.47 $\pm$ 0.07 & 17.47 $\pm$ 1.49 & 26.80 $\pm$ 0.61 & 37.89 $\pm$ 2.37 & 22.44 $\pm$ 3.08 & 27.04 $\pm$ 1.09 & 15.90 $\pm$ 5.63 & 27.82 $\pm$ 3.56 & 20.57 $\pm$ 7.14 & 46.09 $\pm$ 2.24 & 26.89 \\
QuIST$_{10}$           & 43.40 $\pm$ 0.11 & 20.23 $\pm$ 1.14 & 27.08 $\pm$ 0.52 & 38.36 $\pm$ 1.92 & 27.26 $\pm$ 1.78 & 28.32 $\pm$ 1.76 & 23.86 $\pm$ 2.51 & 31.32 $\pm$ 2.38 & 29.98 $\pm$ 3.29 & 47.82 $\pm$ 0.61 & 30.47 \\
QuIST$_{15}$           & 43.08 $\pm$ 0.06 & 19.07 $\pm$ 1.47 & 26.84 $\pm$ 0.49 & 38.79 $\pm$ 3.36 & 27.56 $\pm$ 0.63 & 28.36 $\pm$ 2.63 & 25.14 $\pm$ 1.69 & 30.59 $\pm$ 1.39 & 30.74 $\pm$ 2.02 & 47.71 $\pm$ 0.41 & 30.53 \\ \midrule
GPT-3.5-turbo$_{zero}$ & 33.98            & 21.30            & 27.76            & 35.55            & 24.84            & 31.18            & 18.56            & 27.90            & 17.31            & 41.67            & 27.34 \\
GPT-3.5-turbo$_{10}$   & 37.63            & 21.51            & 29.49            & 39.41            & 26.60            & 32.54            & 22.28            & 30.12            & 23.13            & 44.47            & 29.95 \\
\bottomrule
\end{tabular}}
\caption{\label{tab:appendix:rouge_l} Automatic evaluation results using ROUGE-L and SP-ROUGE.}
\end{table*}

\begin{table*}[ht]
\resizebox{\textwidth}{!}{
\begin{tabular}{l|ccccccccc|c}
\toprule
Model                                        & bn      & de    & fi    & hi    & id    & ko    & sw    & te    & zh    & Avg   \\ \midrule
GPT-3.5-turbo$_{10}$                         & 21.51   & \textbf{29.49} & \textbf{39.41} & \textbf{26.60} & 32.54 & \textbf{22.28} & 30.12 & 23.13 & \textbf{44.47} & 29.95 \\ 
w/ QTC \& Target language Question Exemplars & \textbf{21.97}   & 28.08 & 38.99 & 26.01 & \textbf{34.63} & 20.15 & \textbf{32.43} & \textbf{26.46} & 43.16 & \textbf{30.21}     \\ 
\bottomrule
\end{tabular}}
\caption{\label{tab:appendix/qtc-gpt} Performance of GPT-3.5-turbo$_{10}$ employing the QTC model and question exemplars in target languages.}
\end{table*}

\end{document}